\documentclass[10pt,twocolumn,letterpaper]{article}

\usepackage{iccv}
\usepackage{times}
\usepackage{epsfig}
\usepackage{graphicx}
\usepackage{amsmath}
\usepackage{amssymb}
\usepackage{booktabs}
\usepackage{multirow}
\usepackage{textcomp}
\usepackage{stfloats}

\usepackage{subcaption}
\usepackage{bm}
\usepackage{ifthen}

\usepackage[pagebackref=true,breaklinks=true,letterpaper=true,colorlinks,bookmarks=false]{hyperref}
\usepackage[capitalize]{cleveref}
\crefname{section}{Sec.}{Secs.}
\Crefname{section}{Section}{Sections}
\Crefname{table}{Table}{Tables}
\crefname{table}{Tab.}{Tabs.}
\iccvfinalcopy 

\def\iccvPaperID{7387} 

\ificcvfinal\fi

\begin{document}
	
	\title{\vspace{-0.7cm}GET3D\textminus\textminus: Learning GET3D from Unconstrained Image Collections \vspace{-0.4cm}}

    \author{Fanghua Yu$^{1}$ \quad Xintao Wang$^{2}$ \quad Zheyuan Li$^{1}$ \\
    Yan-Pei Cao$^{2}$ \quad Ying Shan$^{2}$ \quad Chao Dong$^{1, 3}$\protect\footnotemark[2]\\
    $^{1}$SIAT, Chinese Academy of Sciences \quad $^{2}$ARC Lab, Tencent PCG \quad $^{3}$Shanghai AI Lab\\
    {\tt\small fanghuayu96@gmail.com, xintaowang@tencent.com, zheyuanli884886@gmail.com}\\
    {\tt\small \{caoyanpei, yingsshan\}@tencent.com, chao.dong@siat.ac.cn}}

\newboolean{putfigfirst}

\setboolean{putfigfirst}{true}
\ifthenelse{\boolean{putfigfirst}}{
	
	\twocolumn[{%
		\renewcommand\twocolumn[1][]{#1}%
		\vspace{-0.5em}
  
		\maketitle\thispagestyle{empty}
		\begin{center}
			\centering 
			\vspace{-0.3in}
			\includegraphics[width=\linewidth]{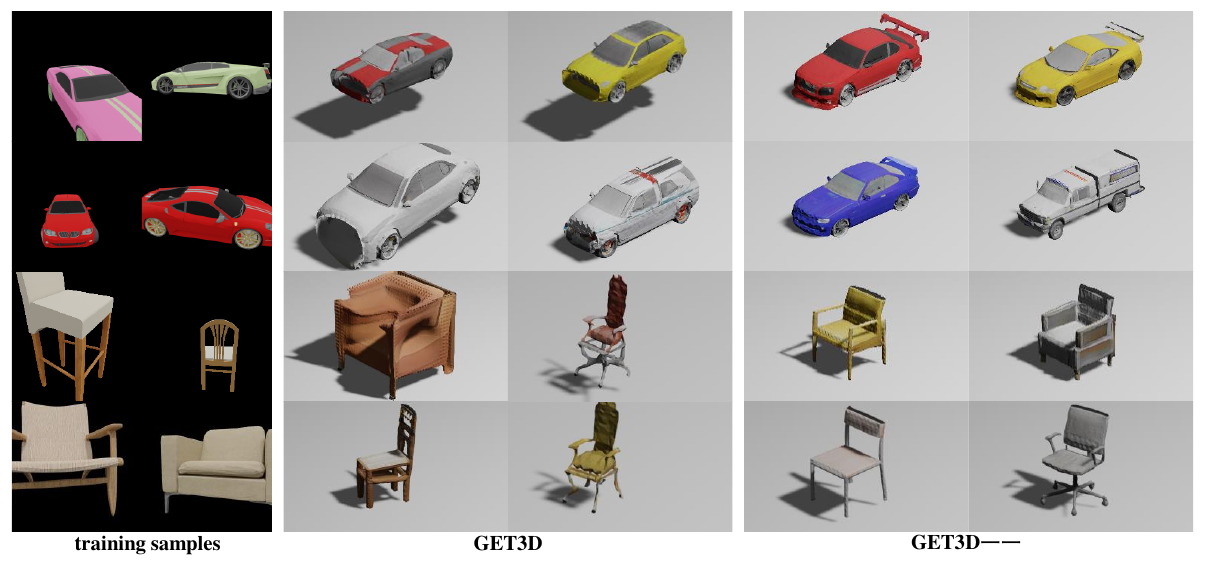}
			\vspace{-0.8cm}
			\captionof{figure}{Examples of 2D training images and the corresponding generated 3D shapes. 
   Using unconstrained 2D image collections with unknown camera pose distributions as training data, GET3D\textminus\textminus~ can learn to generate high-quality textured shapes, while GET3D fails significantly in achieving the same level of performance.
   }
            \vspace{-0.15cm}
			\label{fig:teaser}
		\end{center}%
	}]
}
{
	\maketitle\thispagestyle{empty}
}

\ificcvfinal\thispagestyle{empty}\fi

\newcommand*{\method}{GET3D\textminus\textminus}


\begin{abstract}
\renewcommand{\thefootnote}{\fnsymbol{footnote}}
\footnotetext[2]{Corresponding author (e-mail: chao.dong@siat.ac.cn)}
\renewcommand{\thefootnote}{\arabic{footnote}}
The demand for efficient 3D model generation techniques has grown exponentially, as manual creation of 3D models is time-consuming and requires specialized expertise.
While generative models have shown potential in creating 3D textured shapes from 2D images, their applicability in 3D industries is limited due to the lack of a well-defined camera distribution in real-world scenarios, resulting in low-quality shapes.
To overcome this limitation, we propose \method{}, the first method that directly generates textured 3D shapes from 2D images with unknown pose and scale. \method{} comprises a 3D shape generator and a learnable camera sampler that captures the 6D external changes on the camera.
In addition, We propose a novel training schedule to stably optimize both the shape generator and camera sampler in a unified framework.
By controlling external variations using the learnable camera sampler, our method can generate aligned shapes with clear textures. Extensive experiments demonstrate the efficacy of \method{}, which precisely fits the 6D camera pose distribution and generates high-quality shapes on both synthetic and realistic unconstrained datasets.
\vspace{-0.55cm}
\end{abstract}

\section{Introduction}\label{sec:intro}

With the increasing demand for diverse and high-quality 3D content in various industries, such as gaming, architecture, and social platforms, the ability to efficiently generate 3D models is becoming increasingly important.
However, the manual creation of 3D assets is time-consuming and requires specific expertise and artistic modeling skills.
Generative models for 3D textured shape generation~\cite{graf,voxgraf,pigan,eg3d,get3d} have attracted significant attention in the computer vision community in recent years.

In recent years, researchers have been exploring ways to generate novel views of 3D objects using a set of 2D images. However, voxel-based methods~\cite{graf,voxgraf,pigan,eg3d} commonly used in these works often struggle to produce high-quality shapes that are essential in 3D industries.
To address this limitation, the first work that directly generates textured 3D shapes from 2D images has been proposed, called GET3D \cite{get3d}. This approach employs a surface-based method DMTet \cite{DMTET} to generate 3D shapes and a differentiable renderer \cite{nvdiffrast} to obtain 2D views. The use of a surface-based method and differentiable renderer enables GET3D to produce high-quality 3D shapes that closely resemble the objects in the 2D images, making it a promising direction for future research in 3D shape generation.

However, generating high-quality 3D shapes from \textit{real-world images} remains a challenging task due to several factors, such as shifting lighting, complex material, and uncertain image quality. One specific challenge is the difficulty in obtaining \textit{camera external parameters}, which are often treated as known inputs in most 3D generation methods~\cite{nerf,pixelnerf,mipnerf}. However, this assumption is unrealistic for real images.
Some recent works~\cite{nerf--,selfcnerf, graf} attempt to address this issue by removing the dependence on camera parameters, but they still assume a fixed and known camera distribution, which can lead to poor shapes when there is a large domain gap between the assumed and real camera distribution. To overcome this, some methods~\cite{campari, pof3d} propose to make the camera sampler learnable, but they still have strong assumptions on camera distribution, such as assuming all objects are taken at the center of the camera and the distance between the camera and object is fixed. The existence of these gaps makes it challenging for current 3D generation models to learn high-quality 3D shapes from real-world 2D images.

To address the challenge of generating high-quality 3D textured shapes from 2D images with unconstrained pose, scale, and position, we propose \textbf{\method{}}.
Unlike previous methods, we make the least assumption about the camera external parameter distribution. \method{} consists of a 3D shape generator and a learnable camera sampler that learns 6D external camera pose distribution.
Due to the relative transformation and coupling relationship between the camera and object, joint training of the shape generator\footnote{Throughout this paper, the shape generator generates \textit{textured shapes}, like GET3D\cite{get3d}.} and 6D camera distribution is very unstable, probably leading to the training crash.
To address this, we propose a novel training schedule for staleness, which initializes the shape generator and the camera sampler respectively before joint training.
To guide \method{} generate aligned shapes, we leave all translation and contraction changes as external shifts in the camera by implementing an align loss on the shape generator and a camera compensation on the camera sampler.
Our experiments show that \method{} can exactly fit the 6D camera pose distribution and generate high-quality shapes on unconstrained datasets, as shown in \cref{fig:teaser}.

Our contributions are threefold.
1) \method{} is the first method to directly generate 3D shapes by learning from 6D unconstrained images with unknown camera external parameters.
2) We find that the joint training processes of surface-based 3D generators and 6D camera samplers are fragile, and propose a specific schedule to train these two parts stably.
3) We prove that \method{} can generate aligned high-quality 3D shapes by learning on a complex dataset with a high degree of freedom.

\section{Related Work}\label{sec:related}

\noindent\textbf{3D Generative Models.}
The field of 3D content generation has gained interest following the success of 2D generative models \cite{esser2021taming,huang2022multimodal,sg1,sg2,sg3,adm,ldm,controlnet}. Initially, 3D objects were represented by surface meshes \cite{chen2019learning,pavllo2020convolutional,pavllo2021learning} or point clouds \cite{achlioptas2018learning,yang2019pointflow,zhou20213d}. Later, with advancements in neural rendering \cite{nerf} and implicit neural representation \cite{mescheder2019occupancy,sitzmann2020implicit,chen2019learning,chen2021learning}, methods such as GRAF\cite{graf}, Pi-GAN\cite{pigan}, and EG3D\cite{eg3d} attempted to represent 3D objects in an implicit voxel space.
While these methods can generate novel views, they cannot directly extract 3D objects.
Additionally, voxel-based 3D representation requires more expensive rendering than surface-based methods.
Besides, unlike previous methods, GET3D \cite{get3d} learns geometric and texture features implicitly and transfers them to textured 3D surface meshes.

\noindent\textbf{Camera Sampler in 3D Generative Models.}
NeRF and its variants \cite{pixelnerf,mipnerf,plenoxels} have been widely used for 3D generation, but they heavily rely on accurate camera poses, which are difficult to be obtained in real-world scenarios.
To address this issue, GRAF \cite{graf} and GIRAFFE \cite{giraffe} use distribution constraints instead of pixel-wise constraints to remove the dependence on camera poses, which is also worked in GET3D \cite{get3d}.
However, these methods assume a fixed camera distribution and fail to learn appropriate 3D shapes when the gap between assumed and real camera distributions is large.
CAMPARI \cite{campari} estimates pose distribution but require a closer starting point and manually designed priors, such as uniform or specific Gaussian. PoF3D \cite{pof3d} takes a step forward by abandoning manually designed priors and allowing the generator to learn pose distribution automatically through adversarial training.
However, the concatenation of a learnable camera sampler and surface-based 3D shape generator has not been studied, and previous methods only consider 2D camera poses, which are insufficient for real-world scenarios.

\section{Method}\label{sec:method}

\begin{figure*}[t]
	\begin{center}
		\includegraphics[width=0.9\linewidth]{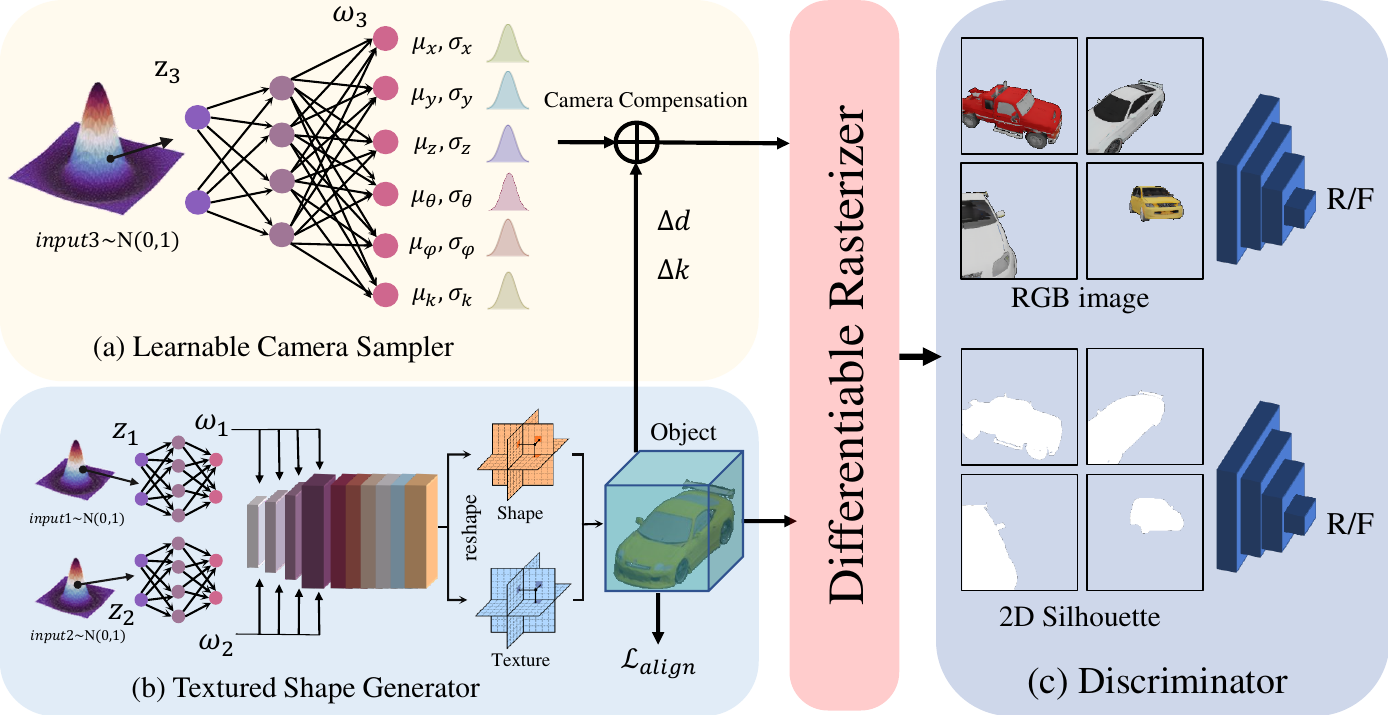}
	\end{center}
	\caption{\textbf{Overview of \method{}}. 
 (a) Learnable Camera Sampler: We firstly sample a camera distribution by the non-linear mapping for ideal Gaussian to the mean and variance values of each camera parameter, then sample a specific 6D camera pose.
 (b) Built upon GET3D~\cite{get3d}, two feature tri-planes are generated by a shared StyleGAN generator \cite{sg2}. DMTet~\cite{DMTET} is applied to modeling surface meshes by querying the shape tri-plane, then texture is queried by the learned surface coordinates from the texture tri-plane.
 (c) Discriminator: After rendered by the differentiable rasterizer -- nvdiffrast~\cite{nvdiffrast}, the output RGB images and its silhouettes are constrained by two independent adversarial losses.
 By applying a learnable camera sampler, \method{} is able to learn 3D shapes on datasets with unknown camera poses and complex distributions.
}
 \vspace{-0.4cm}
\label{fig:framework}
\end{figure*}

In this section, we present our proposed method, \method{}.
First, we introduce the concept of unconstrained images in \cref{subsec:unconstrained}, which is crucial for practical usages.
Then, in \cref{subsec:get3d--}, we describe the overall framework of \method{}.
Finally, in \cref{subsec:schedule}, we discuss the challenges of training a surface-based shape generator and a learnable 6D camera sampler simultaneously, and propose a practical training schedule to achieve stable training.

\subsection{Unconstrained 2D Images}\label{subsec:unconstrained}
Most existing 3D generation methods~\cite{nerf,eg3d,get3d} make strong assumptions about the training data distribution, such as fixed camera-to-object distance, center-focused camera lens, and uniformly sampled camera poses that capture all aspects of the object. However, these assumptions are not easily met in real-world scenarios where objects can be photographed from arbitrary positions, with the object not always centered and not all aspects of the object captured, such as the bottom of a car or the back of a chair. Such strong assumptions result in a significant domain gap between the distribution of training images and real images, making current 3D generative models unsuitable for direct application to real images without adaptation. Learning 3D shapes from unconstrained images is therefore a critical step in developing an ``in the wild" 3D generator.

\subsection{Generative Model of 3D Textured Meshes from Unconstrained Images}\label{subsec:get3d--}
\method{} is a 3D generation framework designed to generate aligned textured surface meshes from unconstrained 2D images, consisting of two main parts: a 3D generator and a learnable camera sampler. The outputs of these blocks are fed into a differentiable rasterizer to render a silhouette and an RGB image, both of which are constrained by an independent 2D discriminator. In addition, to ensure aligned shapes and more accurate estimation of camera distribution, we implement a shape align loss on mesh grids and a camera compensation mechanism on the camera sampler, which decouples the translation and contraction in object and camera.

\noindent\textbf{Representation of Shape and Texture.}
DMTet~\cite{DMTET} is applied to represent shapes in \method{}, which is a differentiable hybrid representation for surface modeling that utilizes a deformable tetrahedral grid and a signed distance field (SDF).
The grid is composed of tetrahedrons, where each tetrahedron is defined by four vertices $\left\{\mathbf{v}_{a_k}, \mathbf{v}_{b_k}, \mathbf{v}_{c_k}, \mathbf{v}_{d_k}\right\}$.
Each vertex is associated with an SDF value $s_i \in \mathbb{R}$ and a deformation value $\Delta \mathbf{v}_i \in \mathbb{R}^3$.
These values are obtained by querying the shape tri-plane with the original coordinate $\mathbf{v}_i \in \mathbb{R}^3$.
The surface of an object is modeled by the set of tetrahedrons whose vertices contain opposite signs in $s_i$.
Additionally, the texture of a learned shape is determined by querying the texture tri-plane with the surface position, which is computed based on the current shape and camera pose.

\noindent\textbf{Feature Generator.}
We build \method{} upon the improved version of GET3D~\cite{get3d}. 
Non-linear mapping networks are utilized to map random Gaussian inputs $z_1$, $z_2$, and $z_3$, to latent vectors $w_1$, $w_2$, and $w_3$.
The generated $w_1$ and $w_2$ vectors are then fed into a StyleGAN generator~\cite{sg2}, which generates both geometry and texture features.
These features are then converted to tri-plane matrices following EG3D~\cite{eg3d}.
Additionally, $w_3$ is used to sample 6D camera poses in the camera sampler, which represents the mean and variance values of camera parameters.

\noindent\textbf{Camera Sampler.}
In \method{}, we consider all 6D camera parameters, including the rotation degree $\theta$, the elevation degree $\varphi$, the object scale $k$, and the 3D displacement $\mathbf{d}$.
We define all possible changes that can be fitted by the camera sampler as extrinsic changes, and design \method{} to avoid these changes from being learned in the shape generator.
To accommodate the requirements of fitting arbitrary camera distributions in real-world scenarios, we model the camera distribution as a joint distribution consisting of one or multiple Gaussian distributions, as shown in the following equation:
\begin{equation}\label{eq:cam_modeling}
	\pi \sim \mathcal{D}_{cam} := \sum_{i=1}^{K} p_i \mathcal{N}(x|\mu_i, \sigma_i^2),
\end{equation}
where $\mathcal{D}_{cam}$ is the distribution of 6D camera parameters. $\mu_i$ and $\sigma_i$ are the mean and variance of the $i$-th Gaussian distribution, respectively. $p_i$ is the normalized probability of sampling camera parameters in this distribution. In practice, $K$, $\mu$, $\sigma$, and $p$ are determined implicitly by the camera mapping network. As shown in \cref{fig:framework}, we map $w_3$ to a 12-D vector that represents the mean and various values of a sampled camera distribution $\mathcal{Z}_i$. Then we sample a specific camera parameter from $\mathcal{Z}_i$ to guide the rendering process of the current shape.

To prevent the shape generator from learning translation and contraction through scaling or adding a bias on $\mathbf{v}_i$, we estimate the translation degree $\Delta \mathbf{d}$ and contraction degree $\Delta k$ of the current shape as follows:
\begin{equation}\label{eq:estimate_pk}
	\begin{split}
		&\Delta \mathbf{d} = \sum_{i=1}^{|T|}\overline{\mathbf{v}}_i/|T|, \\
		&\Delta k = \sum_{i=1}^{|T|}\Vert\overline{\mathbf{v}}_i - \Delta \mathbf{d}\Vert_2/(c_0 |T|),
	\end{split}
\end{equation}
where $|T|$ is the number of tetrahedrons containing surfaces, $\overline{\mathbf{v}}_i$ is the center coordinate of the $i$-th tetrahedron containing a surface, and $c_0$ is a fixed constant that defines the standard scale. We then use camera compensation, which offsets the influence of $\Delta \mathbf{d}$ and $\Delta k$ by adding an opposite bias to the camera sampler, effectively cutting off the influence of object position and scale on the rendered 2D images. To prevent unconstrained translation and contraction leading to extremely large values, we add an align loss on the learned meshes, which is defined as:
\begin{equation}\label{eq:align_loss}
	\mathcal{L}_{align} = \Vert\Delta k - c_0 \Vert_2 + \Vert \Delta \mathbf{d} \Vert_2
\end{equation}

\noindent\textbf{Rendering and Training.}
Acquiring surface meshes, texture grids, and camera parameters, we apply nvdiffrast~\cite{nvdiffrast} to render 3D shapes to 2D RGB images and their corresponding silhouettes.
The rendered images are then fed into two independent discriminators. The overall generation process can be represented by the following equation:
\begin{equation}\label{eq:total_gen}
	{I}_{color}, {I}_{mask} = \textit{R}(\textit{G}_{obj}(z_1, z_2), \textit{G}_{cam}(z_3)),
\end{equation}
where $\textit{R}(i, j)$ denotes rendering from the shape parameter $i$ and the camera parameter $j$. $\textit{G}_{obj}$ and $\textit{G}_{cam}$ are the object and camera generators, respectively.
The total training loss is defined as:
\begin{equation}\label{eq:loss_all}
	L=L(D_{\mathrm{rgb}}, G)+L(D_{\text {mask }}, G)+\mu_1 L_{\mathrm{reg}}+\mu_2 L_{\mathrm{align}},
\end{equation}
where $L_{\mathrm{reg}}$ is the shape regularization loss in GET3D that aims to remove internal floating faces. $\mu_1$ and $\mu_2$ are hyperparameters that control the normalization level.

\subsection{Training Schedule}\label{subsec:schedule}

\begin{figure}[t]
	\begin{center}
		\includegraphics[width=\linewidth]{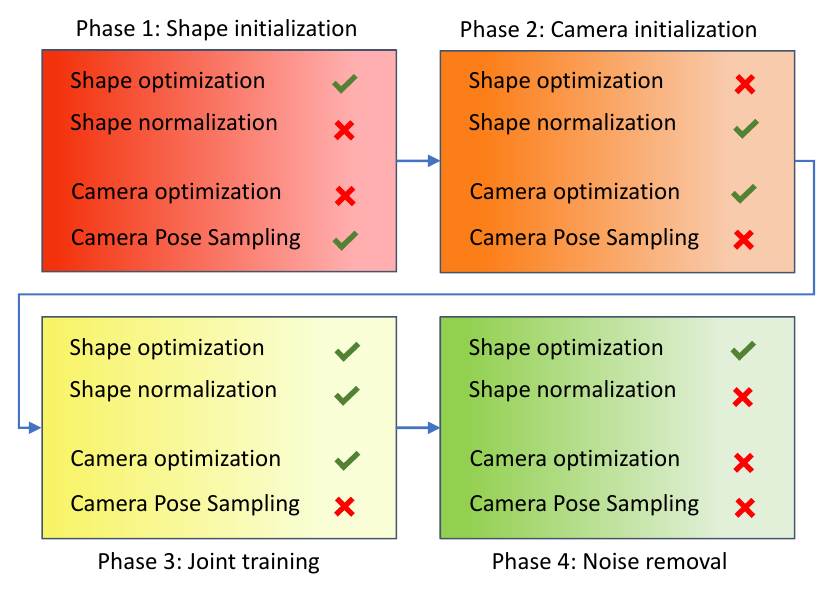}
	\end{center}
	\caption{\textbf{Training schedule of \method{}}. Tick or cross refers to whether this block is activated in the current phase. We initialize the shape generator with a fixed uniform camera sampler and learn specific camera distribution with the learned shapes, followed by joint training of two blocks to reduce estimation errors. This training schedule enables \method{} to converge quickly and stably.}
	\label{fig:schedule}
\end{figure}

In experiments, we observe that joint training of a surface-based 3D generative model and a learnable 6D camera sampler in a single model is challenging to converge and prone to mode collapse. This is mainly due to the assumption space of \method{} being extremely large, which leads to the initialized distribution being hard to converge to the target data distribution. Additionally, the optimization of the shape generator and camera sampler in \method{} is interdependent, making the convergence of each other critical. To address this, we initialize the shape generator and the camera sampler separately before joint training, as illustrated in \cref{fig:schedule}.

\noindent\textbf{Phase1: Initialization of the Shape Generator.}
To obtain a coarse shape, we initially train the shape generator with a fixed camera pose distribution. Since the target camera distribution is unknown, we only sample rotation and elevation degrees from a uniform distribution while fixing the scale and position. During this phase, the 3D generator can learn translation and scaling changes to capture more accurate and clear shapes.

\noindent\textbf{Phase2: Initialization of the Camera Encoder.}
During this phase, we aim to learn a coarse camera pose distribution by calibrating the learned coarse shape with the target shapes. To prevent sudden changes in the camera distribution that may cause significant noise in the shape generator, we keep the shape generator fixed and only train the camera sampler. To learn all external shifts of aligned objects, we implement camera compensation estimated using \cref{eq:estimate_pk}.

\noindent\textbf{Phase3: Joint Training.}
After initializing the two parts of \method{}, we perform joint training to reduce the noise caused by each other.
To constrain the position and scale of learned shapes, we propose the align loss after applying camera compensation.
Since we decouple the function of the shape generator and the camera encoder, and there are no sudden changes after initialization, the joint training is stable in phase 3.

\noindent\textbf{Phase4: Removing External Distraction on Shape Generator.}
Finally, to obtain high-quality 3D shapes, we remove all blocks that introduce noise and focus on training the shape generator alone. One source of noise is caused by the changing distribution of camera parameters, which we address by keeping the camera position fixed. Another source of noise comes from the scale estimation $\Delta k$ in the align loss, which strongly depends on the topological structure of the current shape. For example, the scale estimation on vans and cars may be inconsistent. To overcome this issue, we remove the camera compensation and align loss in the last phase.

In summary, our approach consists of four phases for training a 3D shape generator: initialization of the shape generator with a fixed camera pose distribution, initialization of the camera encoder to learn a coarse camera pose distribution, joint training to reduce noise and constrain the position and scale of learned shapes, and finally, removing external distractions on the shape generator to obtain high-quality 3D shapes. By following this training approach, \method{} can effectively generate accurate 3D shapes while reducing the impact of noise caused by camera parameters and inconsistent scale estimation.


\begin{figure*}[]
    \vspace{0cm}
	\begin{center}
		\includegraphics[width=\linewidth]{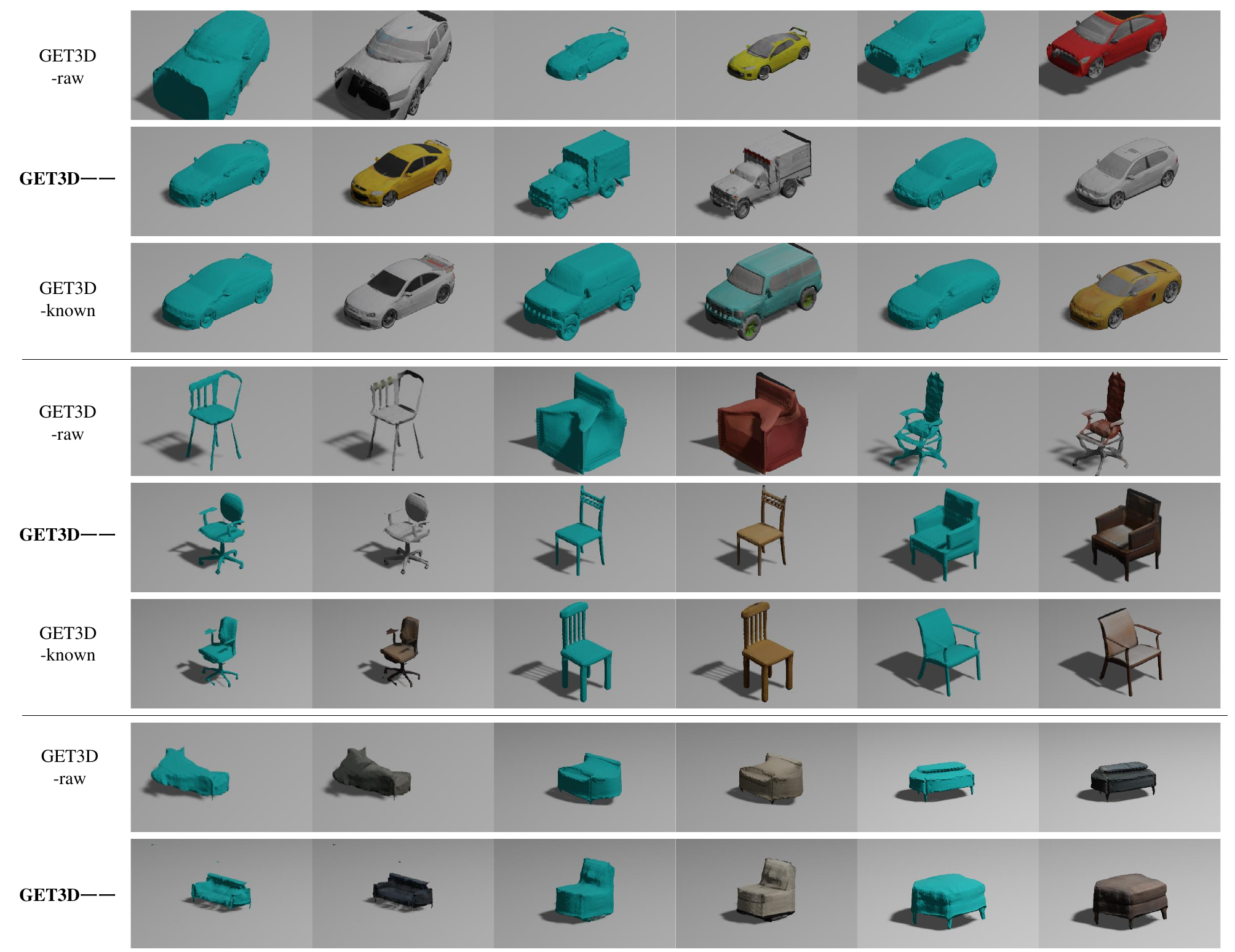}
	\end{center}
    \vspace{-0.2cm}
	\caption{Qualitative results on ShapeNet-Car, ShapeNet-Chair, and ABO-Sofa. For each model, we exhibit 3D surface meshes and textured shapes. All 3D shapes are rendered in Blender.
		\method{} generates high-quality textured 3D shapes and sharp 2D views on both synthetic and real datasets, while the original GET3D with a fixed camera sampler failed to build reasonable shapes. Notice that GET3D-raw fails to generate aligned shapes and we re-scale each object to obtain better viewing of shapes.}
    \vspace{-0.1cm}
	\label{fig:main_visual}
\end{figure*}

\section{Experiments}\label{sec:exp}
In our experiments, we evaluate the performance of \method{} on three datasets: two synthetic and one realistic. The experimental settings are described in detail in \cref{subsec:exp_setup}. We then present the main results of our comparison between GET3D and \method{} in \cref{subsec:main_result}. Finally, we conduct an ablation study to analyze the impact of our training schedules and design choices in \cref{subsec:ablation}. More experimental details and visual comparisons can be found in the supplementary material. 

\subsection{Experimental Setup}\label{subsec:exp_setup}

\noindent\textbf{Datasets.}
We conduct experiments on two categories from the ShapeNet \cite{shapenet} dataset and one category from the Amazon Berkeley Objects (ABO) dataset \cite{abo}. Specifically, we extract high-quality 3D models of cars and chairs from ShapeNet, denoted as ShapeNet-Car and ShapeNet-Chair, and 360$^\circ$ view real images of sofas from the ABO dataset, denoted as ABO-Sofa. To simulate realistic scenarios, we add Gaussian perturbations to the camera distance and object location during the rendering process of the 3D objects in ShapeNet. Additionally, we use a non-uniform Gaussian distribution to sample camera poses with insufficient variance, resulting in sets of sampled 2D images that do not capture the overall perspective of an object. For ABO-Sofa, which consists of 2D images, we directly apply translation and contraction to simulate changes in 3D space. Finally, we obtain three datasets with inconsistent scales, shifted object centers, and non-uniform poses.

\noindent\textbf{Baselines.}
To the best of our knowledge, \method{} is the first work that handles unknown 6D camera distributions during training in a 3D generative model.
Therefore, we mainly compare \method{} with the original GET3D, which uses a fixed and uniform sampler for camera pose sampling (denoted as GET3D-raw).
Additionally, we evaluate the upper bound of GET3D on the current datasets by using a fixed 6D sampler that matches the camera distribution in the rendering process (denoted as GET3D-known).

\noindent\textbf{Metrics.}
We evaluate 3D shape quality using Coverage score (COV) and Minimum Matching Distance (MMD) \cite{achlioptas2018learning}, which measure the mean quality and diversity of generated shapes, respectively. We use Chamfer Distance (CD) and Light Field Distance \cite{chen2017distance} (LFD) as basic functions to measure the similarity of two objects. For texture quality, we use FID on 2D images. Additionally, all objects are normalized to have an identical geometric center and scale before evaluating shapes. In the FID calculation process, we remove noise and sample camera poses uniformly in both fake shape generation and ground-truth rendering. Detailed calculation functions for all metrics can be found in the supplementary material.

\begin{table}[h]
	\centering
	\scriptsize
    \vspace{2mm}
	\begin{tabular}{@{}clccccc@{}}
		\toprule
		\multirow{2}{*}{Category} & \multicolumn{1}{c}{\multirow{2}{*}{Method}} & \multicolumn{2}{c}{COV (\%, $\uparrow$)} & \multicolumn{2}{c}{MMD ($\downarrow$)} & FID ($\downarrow$) \\ \cmidrule(l){3-7} 
		& \multicolumn{1}{c}{} & LFD & CD & LFD & CD & 3D \\ \midrule
		\multirow{3}{*}{Car} & GET3D-raw & \textbf{74.30} & \textbf{56.09} & 1510 & 0.76 & 16.46 \\
		& \method{} & 69.61 & 54.82 & \textbf{1418} & \textbf{0.71} & \textbf{10.58} \\
		& GET3D-known & 52.95 & 42.24 & 1538 & 0.85 & 14.73 \\ \midrule
		\multirow{3}{*}{Chair} & GET3D-raw & 5.96 & 45.89 & 4734 & 5.46 & 41.70 \\
		& \method{} & \textbf{9.50} & \textbf{60.21} & 4022 & \textbf{4.40} & \textbf{32.93} \\
		& GET3D-known & 6.69 & 41.56 & \textbf{3625} & 6.05 & 38.65 \\ \bottomrule
	\end{tabular}
	\caption{Quantitative results on ShapeNet-Car and ShapeNet-Chair. The best result is shown in \textbf{bold}, and MMD-CD scores are multiplied by $10^3$. \method{} outperforms on shape qualities and texture diversity, and achieves even better results than directly sampling from the ground-truth distribution.}
 	\vspace{-10mm}
	\label{tab:main}
\end{table}

\noindent\textbf{Training settings.}
The hyperparameters used in our experiments are based on those used in GET3D \cite{get3d}.
Loss functions are optimized by Adam~\cite{ADAM}, with a learning rate of $0.0002$ and a total batch size of $32$.
All models are trained on a total of $10^7$ images. For \method{}, we proceed to the subsequent training phase at 20\%, 30\%, and 40\% of the total training iterations. 

\subsection{Quantitative and Qualitative Results}\label{subsec:main_result}

\noindent\textbf{Quantitative Results.}
As shown in \cref{tab:main}, \method{} consistently learns more accurate shapes and textures than the original GET3D model on challenging unconstrained image datasets. We find that \method{} outperforms the original GET3D model on both datasets on MMD scores, which indicates that managing the translation and contraction changes in the data by an external camera sampler is beneficial for learning detailed shapes.
Furthermore, we find that GET3D-- achieves even better FID than GET3D-known. We infer that the camera sampler in \method{} is able to capture inner changes in the ground-truth objects, leading to better alignment when querying the textures. 
Note that all models, including \method{} failed to align the object orientation on ShapeNet-Chair, the COV-LFD score is obviously smaller than that in ShapeNet-Car.

\noindent\textbf{Qualitative Comparison.}
In \cref{fig:main_visual}, we showcase the effectiveness of \method{} in generating detailed shapes and textures. \method{} excels in generating intricate shapes, such as the front of cars and the foot of chairs, as demonstrated in the figure.
In contrast, GET3D-raw learns all external changes in the objects, resulting in unaligned shapes with strong artifacts. For instance, chairs with two backs or cars with non-closed surfaces. On the ABO-Sofa dataset, both models face challenges in generating accurate shapes due to the smooth surface and varying lighting conditions of real objects. Moreover, as ABO-sofa is a subset of 360$^\circ$-view images, the variance on elevation is extremely low. This further increases the difficulty of learning reasonable shapes.
However, despite the overall lower quality of the generated shapes, \method{} still outperforms GET3D-raw.

\begin{figure}[b]
    \vspace{-5mm}
	\begin{center}
		\includegraphics[width=\linewidth]{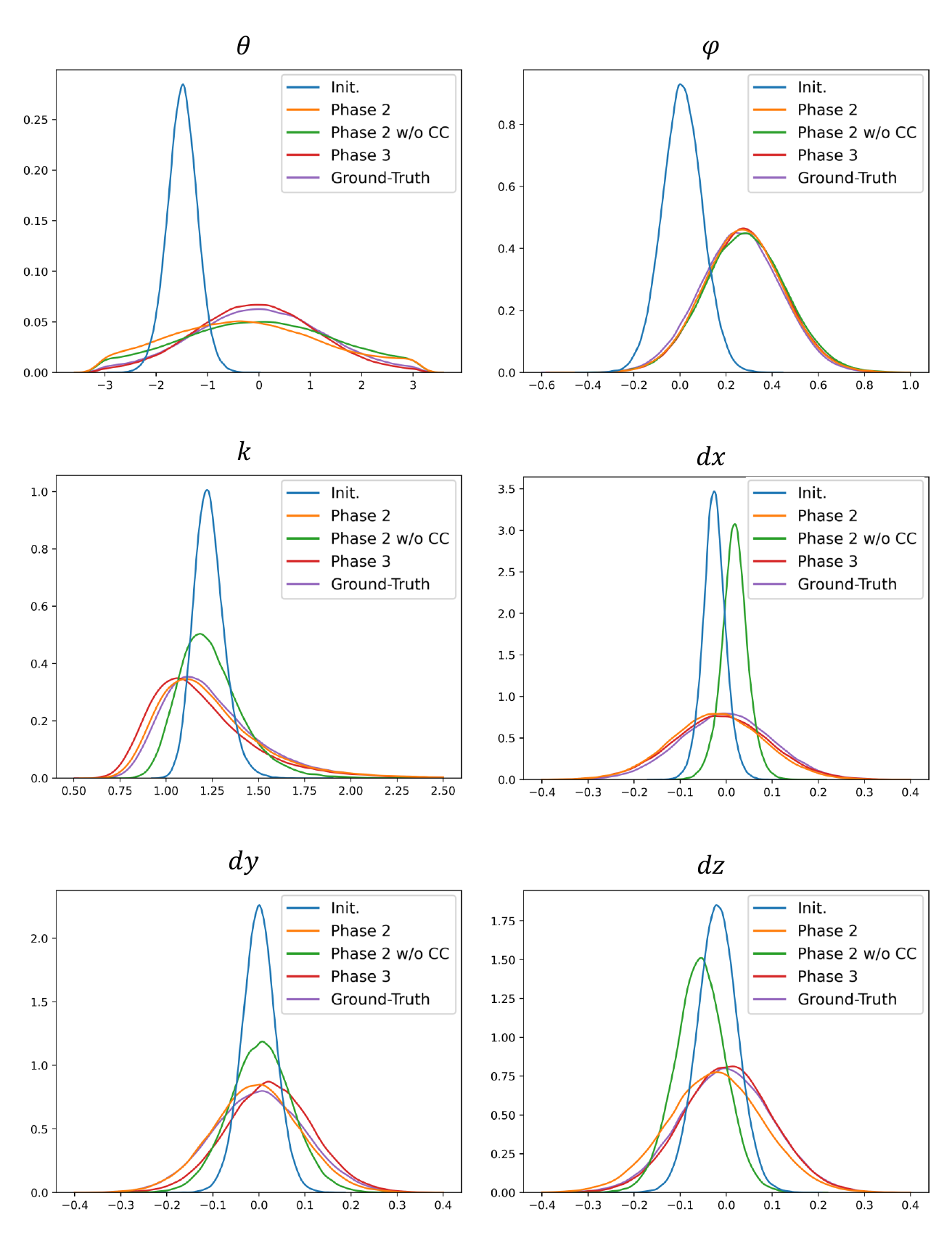}
	\end{center}
    \vspace{-5mm}
	\caption{Visual results during different training phases. Phase 1 generates a coarse and unaligned shape, while phase 3 fixes most artifacts in shapes. Phase 4 focuses on learning details of both shapes and textures. Additionally, removing $L_{\mathrm{align}}$ in phase 3 leads to unconstrained object scale and center position, which further results in generating poor shapes.}
	\label{fig:cam_distribution}
\end{figure}

\subsection{Ablation Study}\label{subsec:ablation}
In this section, we evaluate output models of all phases to show theirs influence on shapes or camera distributions. Further, we make ablations to show the effectiveness of the camera compensation operation and the shape align loss. 

\begin{figure}[b]
	\begin{center}
	\includegraphics[width=\linewidth]{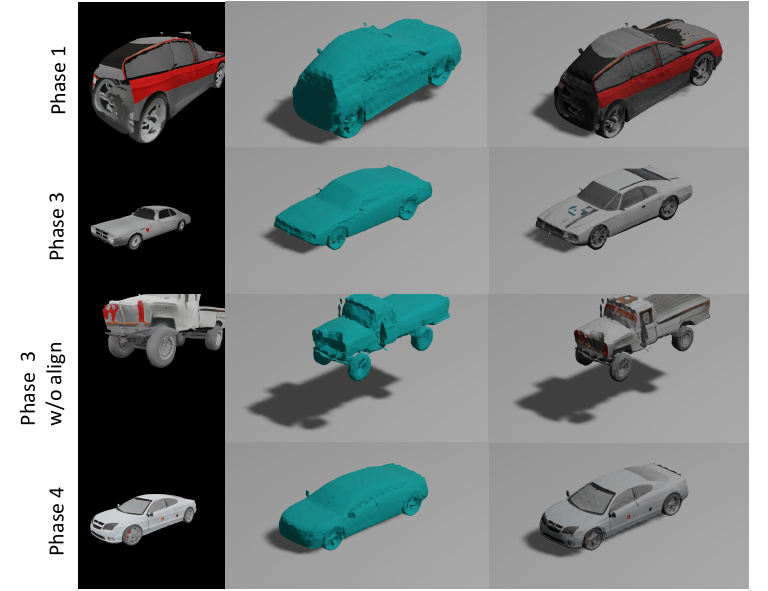}
	\end{center}
	\caption{Visual results of the 6D camera distribution on ShapeNet-Car. The learnable camera sampler in \method{} is converged on a close distribution with ground-truth. Phase 3 mainly modifies the distribution of the rotation degree. CC refers to 'camera compensation', which benefits to predict a more accurate variance.}
	\label{fig:ablation}
\end{figure}

\noindent\textbf{Object Learning Phases.}
For all phases that optimize the shape generator, we conduct both quantitative and qualitative analyses to assess the quality of the generated shapes. 
Our results, presented in \cref{tab:ablation}, show that when initialized with a fixed and non-conformity camera sampler, phase 1 is only able to learn coarse shapes. However, the quality of generated shapes is significantly improved in phase 3, which benefits from the more accurate estimation in the camera sampler.
Although there is no significant improvement in shape quality beyond phase 3, we find that phase 4 generates more detailed textures, as indicated by lower FID scores. In \cref{fig:ablation}, we observe that the shapes learned in phase 1 are able to capture the abstract outline of the object, which is sufficient to learn a coarse camera distribution. As phase 3 addresses most of the artifacts in shape, it still produces noisy textures. This issue is resolved in phase 4.

\begin{table}[t]
    \vspace{2mm}
	\centering
	\scriptsize
	\begin{tabular}{@{}lccccc@{}}
		\toprule
		\multicolumn{1}{c}{\multirow{2}{*}{Schedule}} & \multicolumn{2}{c}{COV (\%, $\uparrow$)} & \multicolumn{2}{c}{MMD ($\downarrow$)} & FID ($\downarrow$) \\ \cmidrule(l){2-6} 
		\multicolumn{1}{c}{} & LFD & CD & LFD & CD & 3D \\ \midrule
		Phase 1 & \textbf{74.77} & 54.69 & 1536 & 0.83 & 19.6 \\
		Phase 3 & 63.39 & 48.93 & 1487 & 0.77 & 15.3 \\
		Phase 3 w/o $L_{\mathrm{align}}$ & 55.15 & 45.05 & 1495 & 0.84 & 78.62 \\
		Phase 4 & 69.61 & \textbf{54.82} & \textbf{1418} & \textbf{0.71} & \textbf{10.58} \\ \bottomrule
	\end{tabular}
	\caption{Ablation results on ShapeNet-Car. The best result is shown in \textbf{bold}. \method{} benefits from all designed phases and $L_{\mathrm{align}}$ is vital for stable training.}
    \vspace{-4mm}
	\label{tab:ablation}
\end{table}

\begin{figure}[b]
    \vspace{-2mm}
	\begin{center}
		\includegraphics[width=\linewidth]{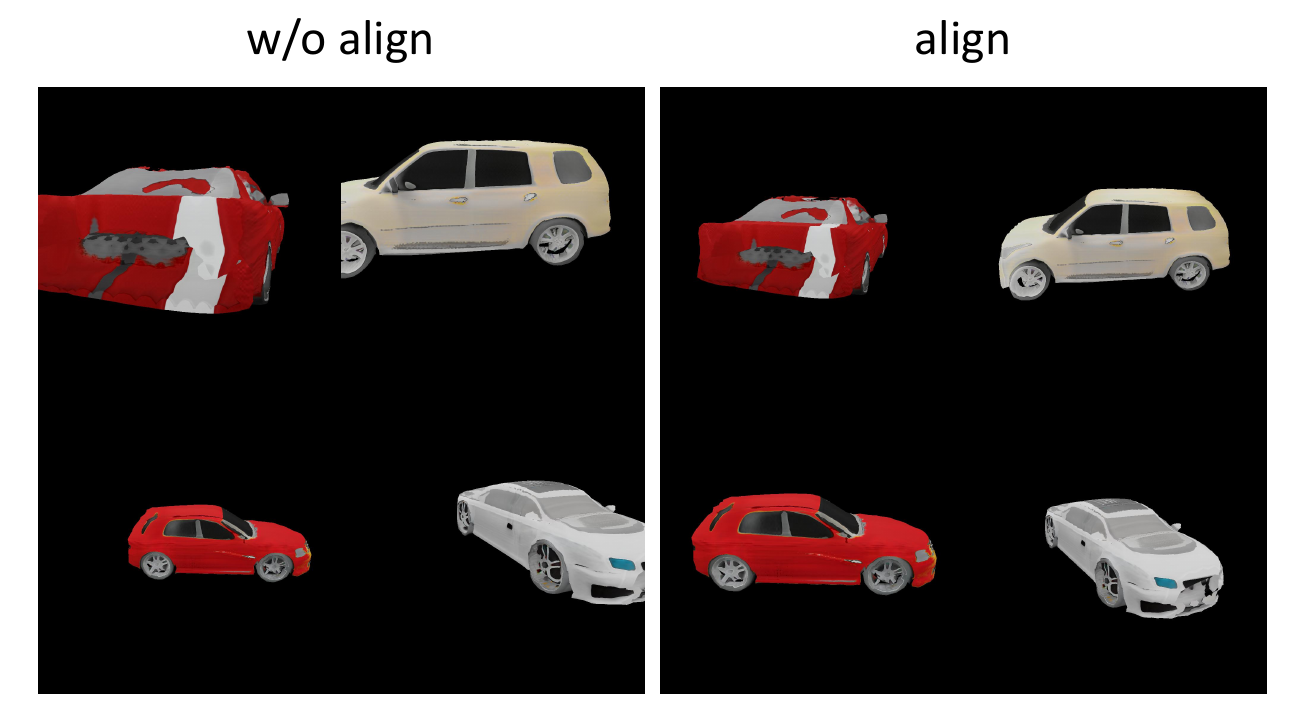}
	\end{center}
    \vspace{-4mm}
	\caption{Visualization of the camera compensation influences. As we remove most of internal translation and contraction changes by camera compensation, we can predict more accurate external changes in the camera sampler.}
    \vspace{2mm}
	\label{fig:cam_align}
\end{figure}

\noindent\textbf{Camera Learning Phases.}
In \cref{fig:cam_distribution}, we present the visualization of the distribution of all components in the camera sampler.
To simulate the realistic scenario, the initialized mean values of rotation degree $\theta$, elevation degree $\varphi$ and camera distance $k$ are designed to keep away from their ground-truth values.
The camera distribution is initialized with a small variance, which helps to quickly converge to the mean value of each distribution.
In phase 2, the camera distribution is already closer to the ground-truth distribution on most components, but there are issues in fitting the rotation degree due to the rotation changes learned in objects.
However, as the object and camera distributions are jointly optimized, this issue is resolved.
Overall, our results indicate that the camera sampler in \method{} effectively learns external changes in all dimensions.

\textbf{Camera Compensation.}
In \cref{fig:cam_align}, we demonstrate the effectiveness of the camera compensation technique in removing internal translation and contraction changes in objects, enabling the camera sampler to learn all possible changes.
This is further supported by our findings in \cref{fig:cam_distribution}, where we observe that removing the camera compensation prevents the camera sampler from learning large variances in most components.
Besides, given a surface meshes, translation or contraction changes can be achieved by adding a bias or multiplying a scale coefficient on its vertex coordinates. However, performing rotation changes is more complex, which requires matrix multiplication. Consequently, we do not perform compensations on internal rotation changes, and it will be removed automatically in phase 3. 
Overall, our results suggest that the camera compensation technique is crucial for enabling the camera sampler to effectively capture external changes in objects.

\noindent\textbf{Shape Align Loss.}
Our results in \cref{tab:ablation} and \cref{fig:ablation} demonstrate that without the use of $L_{\mathrm{align}}$, the scale and geometric center of objects are unconstrained, resulting in poor quality generated objects with regards to both shapes and textures. This is due to the fact that both vertex deformation and SDF values of shapes, as well as RGB values of textures, are queried from tri-planes using coordinates. Any changing biases or scales in these coordinates can lead to poor shape quality or even shape collapse. Therefore, $L_{\mathrm{align}}$ is critical for ensuring accurate shape and texture generation.
As discussed in \cref{subsec:schedule}, applying $L_{\mathrm{align}}$ can introduce noise during the learning of objects, as it is coupled with shapes. To address this, we introduce phase 4 and set a relatively small weight to $L_{\mathrm{align}}$.
In addition, when facing large changes in the object shapes, we usually use the maximum distance of any two points on the object to estimate the object scale. However, applying such a loss function calculated with few feature points can lead to training instability.

\section{Conclusion}\label{sec:conc}
In this paper, we introduce \method{}, a 3D generative model that can learn textured shapes from 2D images with an unknown pose, scale, and position. By utilizing a learnable camera sampler to control external variations, \method{} is able to generate shapes that are aligned and have clear textures. We address the issue of stably training a shape generator and a camera sampler in a unified framework, by proposing a feasible training schedule. This schedule initializes the two parts separately before joint training. We also propose a compensation on the camera and a constraint on the object to guide the shape generator in generating more aligned shapes. Extensive experiments demonstrate the effectiveness of \method{}, which precisely fits the 6D camera pose distribution and generates high-quality shapes on both synthetic and realistic unconstrained datasets. \method{} proves the potential to develop a real "in the wild" 3D generative model by managing all possible external changes in cameras.

\noindent\textbf{Limitation.}
There are several limitations that should be noted:
1) The camera sampling schedule in phase 1 needs to have some overlap with the ground-truth camera distribution, or else we cannot learn an abstract shape to perform the next phase. This means that the camera distribution needs to be carefully designed to ensure that it covers a wide range of possible viewpoints.
2) The noise introduced by the $L_{\mathrm{align}}$ loss is tolerable when the learning shapes are similar, but it may lead to failure results when the objects vary greatly in shape. Future work could explore more robust ways to align objects and cameras.
3) In this paper, we assume that all camera components are independent and its ground-truth distribution is designed to be a single Gaussian. Further work could explore more complex camera distributions and their impact on our approach.

{\small
	\bibliographystyle{ieee_fullname}
	\bibliography{egbib}
}

\clearpage
\appendix

\noindent\textbf{\large{Appendix}} \label{appendix}

\begin{table*}[hb]
	\centering
    \vspace{0mm}
	\begin{tabular}{@{}lcccccc@{}}
\toprule
Dataset & Shapes & Views per shape & Rotation Angle & Elevation Angle & Objcet Scale & Objcet Position \\ \midrule
ShapeNet-Car & 7497 & 24 & $\mathcal{N}(\pi, \frac{2}{5}\pi)$ & $\mathcal{N}(\frac{5}{6}\pi, \frac{1}{6}\pi)$ & $\mathcal{N}(1.2, 0.2)$ & $\mathcal{N}(0, 0.2)$ \\
ShapeNet-Chair & 6778 & 24 & $\mathcal{N}(\pi, \frac{2}{5}\pi)$ & $\mathcal{N}(\frac{5}{6}\pi, \frac{1}{6}\pi)$ & $\mathcal{N}(1.2, 0.2)$ & $\mathcal{N}(0, 0.2)$ \\
ABO-Sofa & 959 & 64 & $\mathcal{N}(\pi, \frac{2}{5}\pi)^*$ & Fixed \& Unknown & $\mathcal{N}(1024, 256)^*$ & $\mathcal{N}(0, 128)^*$ \\ \bottomrule
\end{tabular}
\vspace{-2mm}
	\caption{Experimental settings of each datasets. We utilize Blender to sample 2D images from 3D objects for ShapeNet-Car and ShapeNet-Sofa, using all available camera parameters. As for ABO-Sofa, we begin by sampling a specific rotation degree and selecting the nearest 2D image with the corresponding rotation angle. We then adjust the object's shape and position by applying planar scaling and translation.}
 	\vspace{-2mm}
	\label{tab:dataset}
\end{table*}

\vspace{2mm}
Due to the lack of space in the main paper, we provide more details of the proposed \method{} in the supplementary file.
We outline the experimental details of \method{} in \cref{sec:exp}. This includes information on the datasets used, as well as details regarding our selection of training hyper-parameters and calculation metrics. In \cref{sec:cam_ab}, we present our ablation results on external camera parameters to better understand the impact of learning each parameter group. We also test \method{} in a sparse case in \cref{sec:sparse}, which is a common occurrence in real-world scenarios. Furthermore, we discuss the challenges of ABO-Sofa in \cref{sec:abo}. Finally, to provide a more visually appealing representation of our approach, we include snapshots of training images and visual results in \cref{sec:visual results}.

\section{Experimental Details}\label{sec:exp}

\subsection{Datasets}

\textbf{ShapeNet-Car and ShapeNet-Sofa \cite{shapenet}.}
In our study, we utilize ShapeNet-Car and ShapeNet-Sofa, which contain 7497 and 6776 textured 3D objects, respectively. We employee Blender to generate 2D images by rendering these 3D objects. Each object is sampled to produce 24 images. To replicate real-life conditions, we assume that all camera external parameters followed a Gaussian distribution. In particular, the rotation angle follows a Gaussian distribution with minor variation to ensure that the back of the object is inadequately sampled. For all settings in the rendering processes, refer to \cref{tab:dataset}.

\textbf{ABO-Sofa \cite{abo}.}
ABO-Sofa contains 959 objects, each of which has 72 $360^{\circ}$-viewing images differing by $5^{\circ}$. To handle the lack of 3D object, we perform object scaling and translation in a plain manner. We first apply a center crop with variable scaling and random translation to an input image with a resolution of $1024 \times 1024$, then resize it to the same resolution. In addition, we manually add a mask channel to each image, as the ABO dataset does not provide one. 

In all, we obtain $180k$, $123k$, and $61k$ images in ShapeNet-Car, ShapeNet-Sofa, and ABO-Sofa, respectively. We split the dataset into training (70$\%$), validation (10$\%$), and testing (20$\%$) sets.

\subsection{Training Details}

Our approach, \method{}, builds on the official PyTorch implementation of StyleGAN2 \cite{sg2}. We adopt the same training configuration as StyleGAN2, utilizing techniques such as minibatch standard deviation, exponential moving average, non-saturating logistic loss, and R1 regularization. We train our model from scratch with 2D discriminators, without relying on progressive training or pretrained checkpoints. Our hyperparameters are largely adopted from StyleGAN2, including the Adam optimizer with a learning rate of 0.002 and $\beta$ of 0.9. R1 regularization is applied to the discriminators every 16 training steps with a weight of 80 for ShapeNet-Car, 400 for ShapeNet-Chair, and 400 for ABO-Sofa. Additionally, we use lazy regularization for R1 regularization.
For SDF regularization, we set the hyperparameter $\mu_1$ to 0.01.
$\mu_2$ is set to 0.1 to constrain object scales and positions. We train our model on 8 A6000 GPUs with a batch size of 32 for all experiments. A single model takes approximately 2 days to converge.

In regards to our camera sampler, \cref{tab:cam_distribution} displays the distributions for the fixed sampler used in Phase 1, as well as the initialized distribution for our proposed learnable sampler in Phase 2. Both distributions intentionally deviate from the target distribution, which seeks to replicate real-world scenarios with unknown distributions.

\begin{table}[t]
	\centering
    \small
    \vspace{0mm}
	\begin{tabular}{@{}lccc@{}}
\toprule
 & Fixed & Initialized & Target \\ \midrule
Rotation Angle & $\mathcal{U}(0, 2/pi)$ & $\mathcal{N}(0, \frac{1}{10}\pi)$ & $\mathcal{N}(\pi, \frac{2}{5}\pi)$ \\
Elevation Angle & $\mathcal{U}( \frac{1}{3}\pi, \frac{1}{2}\pi)$ & $\mathcal{N}(\pi, \frac{1}{36}\pi)$ & $\mathcal{N}(\frac{5}{6}\pi, \frac{1}{6}\pi)$ \\
Objcet Scale & $1.2$ & $\mathcal{N}(1.2, 0.06)$ & $\mathcal{N}(1.2, 0.2)$ \\
Objcet Position & $0$ & $\mathcal{N}(0, 0.06)$ & $\mathcal{N}(0, 0.2)$ \\ \bottomrule
\end{tabular}
    \vspace{-2mm}
	\caption{Comparisons of camera parameters distributions. We maintain a deviation between the fixed and initialized camera distributions and the target distribution.}
 	\vspace{-5mm}
	\label{tab:cam_distribution}
\end{table}

\subsection{Metrics}
To evaluate the quality of both texture and geometry, we compare generated shapes $S_g$ to the reference ones $S_r$. Notice that the data scale of $S_g$ is 5 times of $S_r$.

\textbf{Calculating Distance of Given Shapes.}
We use Chamfer Distance $d_{CD}$ and Light Field Distance $d_{LFD}$ \cite{chen2017distance} to measure the similarities of the shapes.
Let $X \in S_g$ denotes a generated shape and $Y \in S_r$ denotes a reference one. $x$ and $y$ are sampled points from the surface of $X$ and $Y$, respectively. Here we sample 2048 points in all experiments.
The $d_{CD}$ can be calculated as:
\begin{equation}
	\begin{split}
d_{\mathrm{CD}}\left(X, Y\right)=&\sum_{\mathbf{x} \in X} \min _{\mathbf{y} \in Y}\|\mathbf{x}-\mathbf{y}\|_2^2+\\
                            &\sum_{\mathbf{y} \in Y} \min _{\mathbf{x} \in X}\|\mathbf{x}-\mathbf{y}\|_2^2 .
	\end{split}
\end{equation}

$d_{LFD}$ renders the shapes $X$ and $Y$ from a set of selected viewpoints, encodes the rendered images using Zernike moments and Fourier descriptors, and computes the similarity over these encodings. Formal definition is available in \cite{chen2017distance}.

\textbf{Evaluating the Geometry.}
Coverage and Minimum Matching Distance are then proposed to evaluation geometric similarities of two distribution.
Coverage (COV) measures the fraction of shapes in $S_r$ that are matched to at least one of the shapes in $S_g$:
\begin{small}
\begin{equation}\label{eq:cov}
COV(S_g, S_r)=\frac{|\{\operatorname{argmin}_{X \in S_r} D(X, Y) \mid Y \in S_g\}|}{|S_r|}.
\end{equation}
\end{small}

Minimum Matching Distance (MMD) measures the mean quality of the generated shapes:
\begin{small}
\begin{equation}\label{eq:mmd}
\operatorname{MMD}\left(S_g, S_r\right)=\frac{1}{\left|S_r\right|} \sum_{X \in S_r} \min _{Y \in S_g} D(X, Y).
\end{equation}
\end{small}

In \cref{eq:cov,eq:mmd}, $D$ can be either $d_{CD}$ or $d_{LFD}$.

\textbf{Evaluating the Texture.}
We utilize the Fréchet Inception Distance (FID) on rendered 2D views to assess the quality of our generated textures. Specifically, we render 50k views of the generated shapes, with one view per shape, randomly sampled from the predefined camera distribution. We also render reference objects from the test set using the same camera distribution.
During evaluating processes of FIDs, we use the fixed distribution in \cref{tab:cam_distribution}.
By utilizing a camera distribution with uniforming rotation angles, we ensure an equal assessment of the textures in each position.

\section{Ablation of External Camera Parameters}\label{sec:cam_ab}
We split the 6D camera parameters into three groups, namely rotation ($\theta$ and $\phi$), scaling ($k$), and translation ($dx$, $dy$, and $dz$), to examine the influence of learning the distribution of each parameter group. We treated one group as pre-defined parameters in all phases and learned the remaining two groups.

\begin{figure}[]
    \vspace{0mm}
	\begin{center}
		\includegraphics[width=\linewidth]{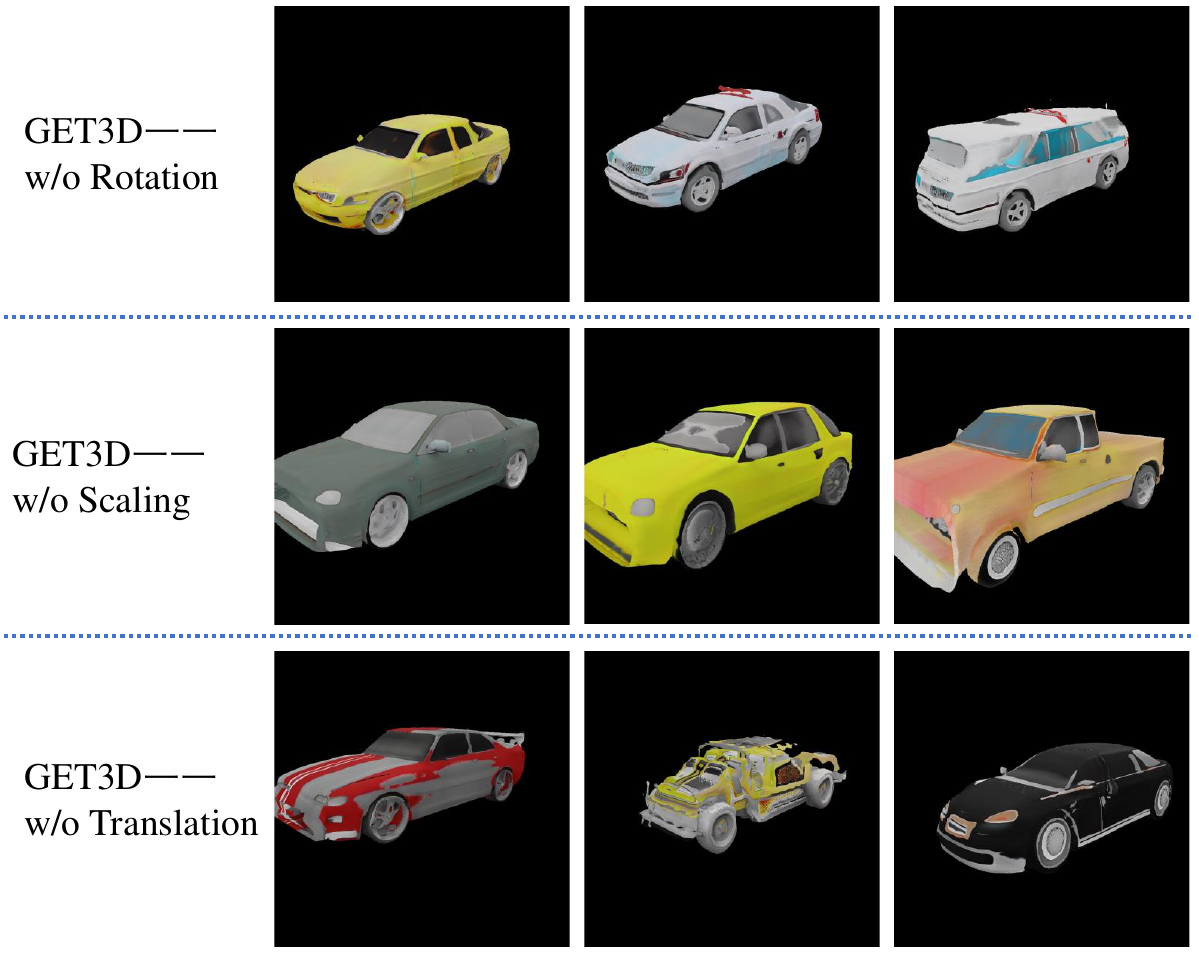}
	\end{center}
    \vspace{-4mm}
	\caption{Visual ablations of camera parameter groups. 
 Unlearnable rotation parameters lead to the generation of cars with double heads; unlearnable scaling parameter causes out-of-bound shapes; and unlearnable position parameters result in poor quality textures.}
	\label{fig:cam_ablation}
\end{figure}

The ablation results, presented in \cref{fig:cam_ablation} and \cref{tab:cam_ablation}, demonstrate the importance of learning the distribution of all three camera parameter groups in generating accurate shapes. Not learning the rotation parameters leads to incorrect semantic information in the generated shapes, while fixing the scaling and translation parameters introduces noise in querying shape and texture information on tri-plane feature maps, resulting in poor quality shapes and textures.

\begin{table}[]
	\centering
    \small
    \vspace{0mm}
	\begin{tabular}{@{}lccccc@{}}
\toprule
\multirow{2}{*}{Method} & \multicolumn{2}{c}{COV (\%, $\uparrow$)} & \multicolumn{2}{c}{MMD ($\downarrow$)} & FID ($\downarrow$) \\ \cmidrule(l){2-6} 
 & LFD & CD & LFD & CD & 3D \\ \midrule
GET3D-raw & 74.30 & 56.09 & 1510 & 0.76 & 16.46 \\
w/o Rotation & 64.86 & 48.86 & 1522 & 0.84 & 12.89 \\
w/o Scaling & 73.29 & 57.10 & 1471 & 0.75 & 17.00 \\
w/o Translation & \textbf{74.36} & \textbf{58.77} & 1474 & 0.75 & 16.57 \\
\method{} & 69.61 & 54.82 & \textbf{1418} & \textbf{0.71} & \textbf{10.58} \\ \bottomrule
\end{tabular}
	\caption{Ablation results on ShapeNet-Car. 'W/o' denotes leaving the group of camera parameters fixed as Phase 1. Our results demonstrate that fitting all three groups is crucial for achieving high-quality textured shapes.}
 	\vspace{-4mm}
	\label{tab:cam_ablation}
\end{table}

\begin{figure}[]
	\begin{center}
		\includegraphics[width=\linewidth]{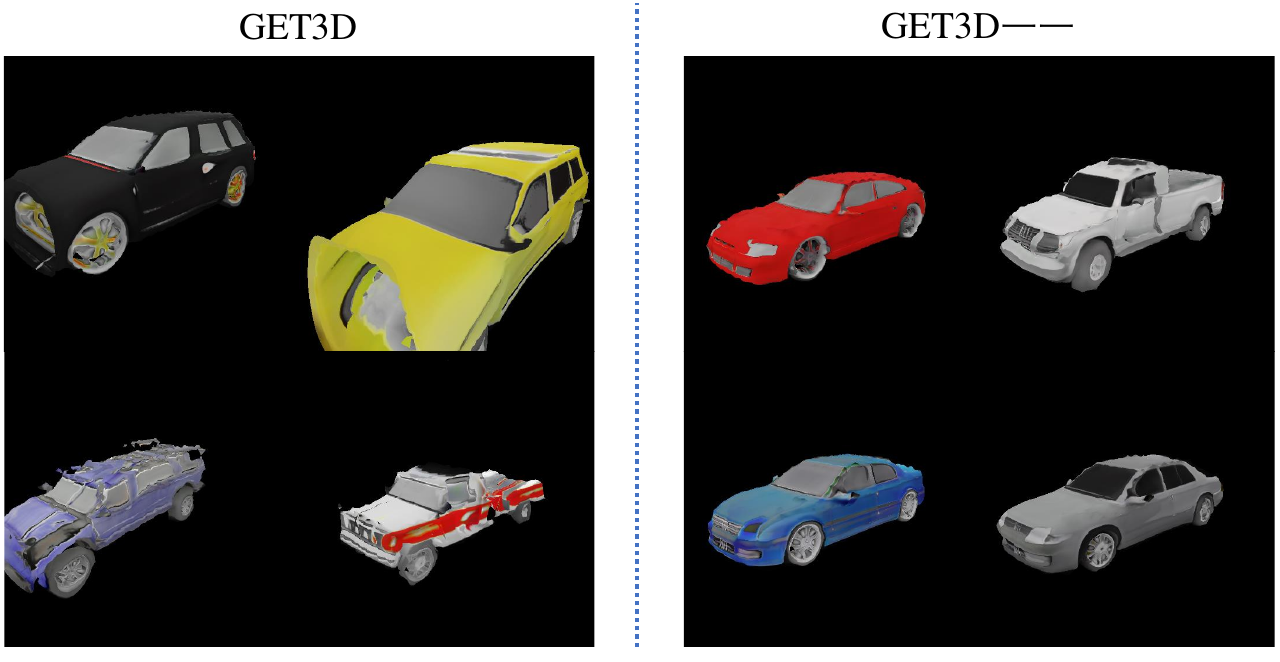}
	\end{center}
    \vspace{-4mm}
	\caption{Visual comparison of GET3D and \method{} on sparse training data. Despite the challenging problem of limited sampling data per object, \method{} is capable of producing reasonable shapes. In contrast, GET3D fails to achieve satisfactory results.}
    \vspace{0mm}
	\label{fig:sparse}
\end{figure}

\begin{figure}[]
	\begin{center}
		\includegraphics[width=\linewidth]{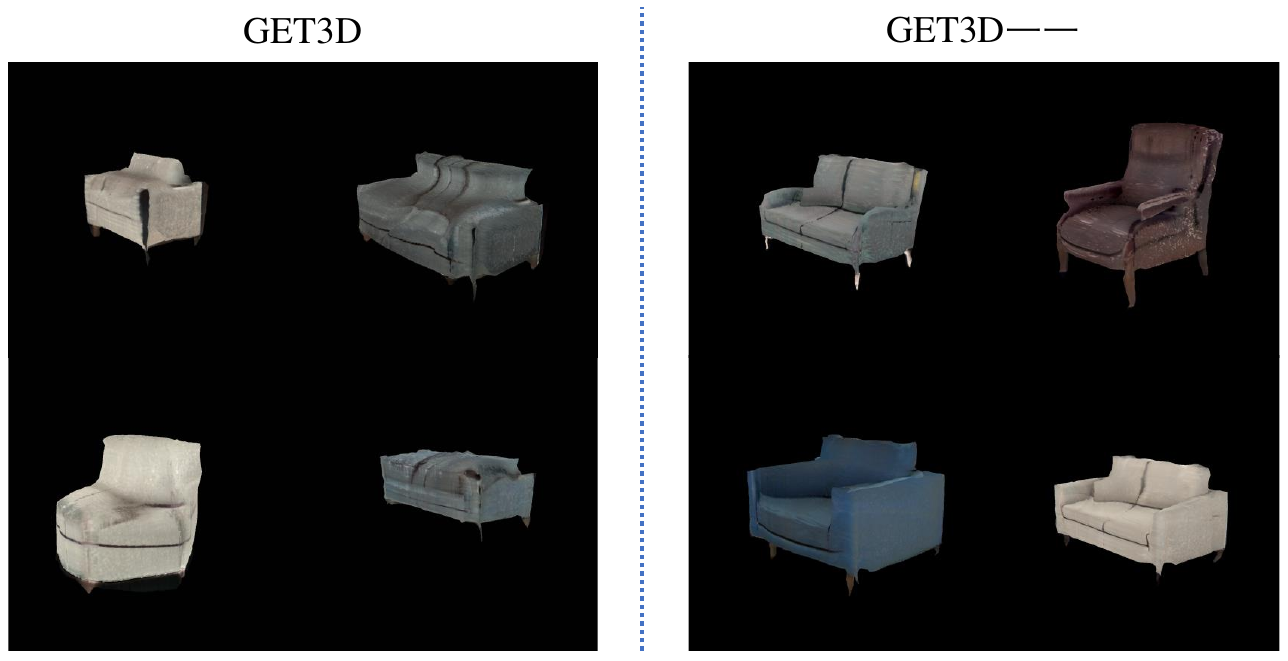}
	\end{center}
    \vspace{-4mm}
	\caption{2D Visual comparison of GET3D and \method{} on ABO-Sofa. \method{} generates accurate contours.}
	\label{fig:sofa_compare}
\end{figure}

\section{Model Stability on Sparse View Data}\label{sec:sparse}
To test the stability of our model under extremely challenging conditions, we conducted experiments with a sparse dataset where the number of samples per object was decreased from 24 to 3. This scenario is common in real-world applications and poses a significant challenge to current 3D generation models. We also reduced the resolution of input images, tetrahedral meshes, tri-planes, and channel widths of all layers in both the generator and discriminator by half.

As shown in \cref{fig:sparse}, unlike GET3D, our proposed method, \method{}, is capable of generating reasonable shapes even when trained with a sparse dataset.

\section{Discussions}\label{sec:abo}

\textbf{Failure cases on ABO-Sofa.}
As shown in \cref{fig:sofa}, \method{} struggles to generate accurate inner structures of sofas, mainly due to insufficient supervision in the current shape generation pipeline. Learning reasonable shapes from ABO-Sofa is even more challenging than from ShapeNet-Car, since objects in ABO-Sofa are typically colored in a consistent color with blurred edges, and their inner structures are opaque. Additionally, all objects in ABO-Sofa are viewed from a limited elevation angle, so a generated object with an incorrect inner structure can still produce reasonable 2D images. When inner shapes have less influence on 2D views and silhouettes, learning accurate shapes becomes a challenge for current 3D generative models. This remaining problem can be addressed by incorporating additional depth information in supervision in future works. While \method{} focuses on learning camera distribution, it generates more accurate outer structures, as shown in \cref{fig:sofa_compare}.

\begin{figure}[]
    \vspace{-2mm}
	\begin{center}
		\includegraphics[width=\linewidth]{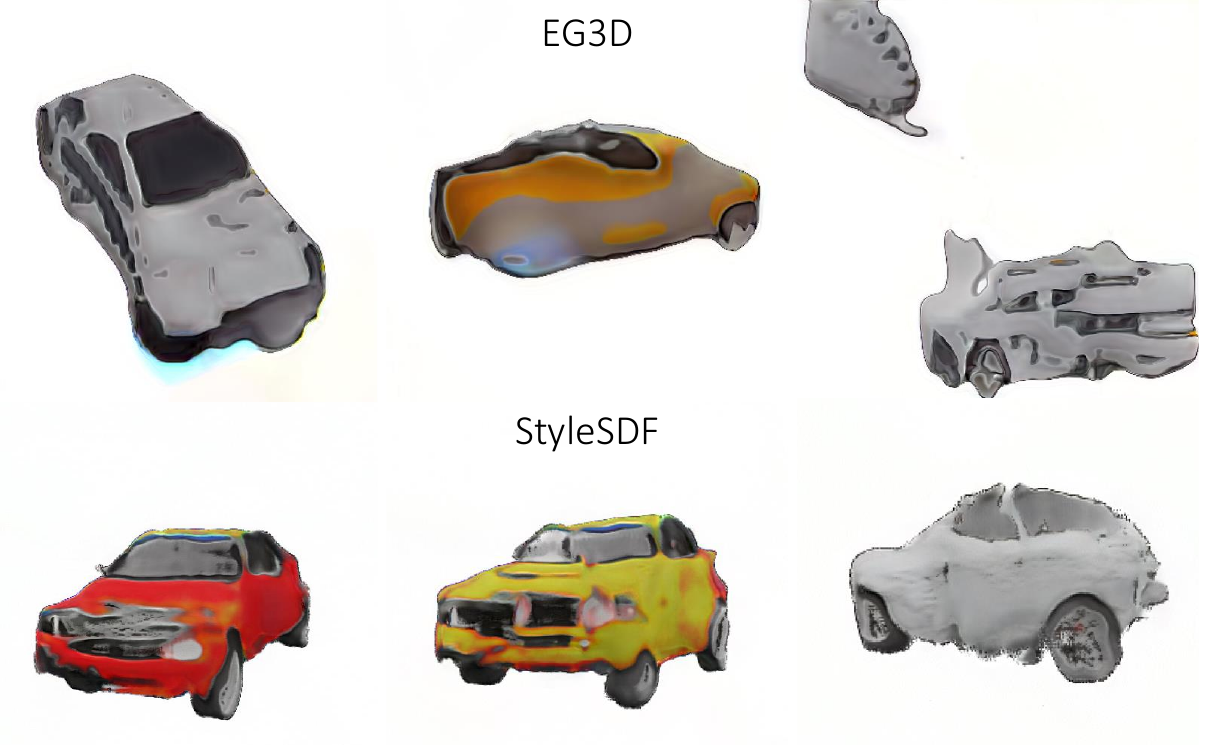}
	\end{center}
	\caption{2D Visual results of EG3D\cite{eg3d} and StyleSDF\cite{or2022stylesdf} on ShapeNet-Car.}
	\label{fig:failed_baseline}
\end{figure}

\textbf{Insufficient baseline.}
The superiority of GET3D over EG3D and StyleSDF has been established in the GET3D paper. 
We choose GET3D as our baseline model because it is the only one capable of successfully generating the coarse shape in our experimental setting. Implicit field-based methods such as NeRF, struggle with inadequate regularization to constrain the overall structure, making it challenging to handle object translation and scaling in our unknown camera setting. 
As illustrated in \cref{fig:failed_baseline}, EG3D and StyleSDF exhibit inferior performance.
Besides, they both fail to extract meaningful mesh surface, making it hard to calculate metrics like CD and LFD.

\begin{figure}[]
    \vspace{-2mm}
	\begin{center}
		\includegraphics[width=\linewidth]{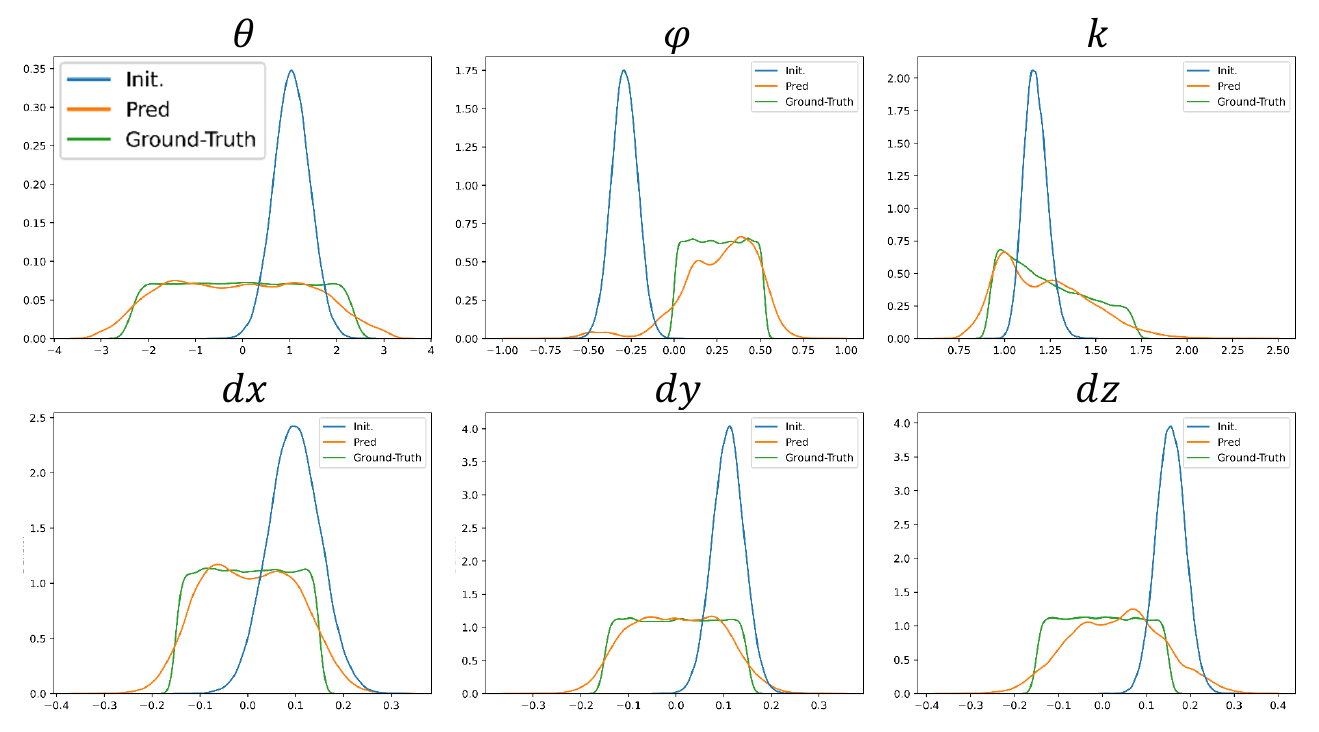}
	\end{center}
    \vspace{-4mm}
	\caption{Visualizations of camera poses under the uniform distribution.}
    \vspace{-6mm}
	\label{fig:uniform_cam}
\end{figure}

\textbf{Synthetic camera distribution.}
Our camera encoder employs a joint distribution consisting of independent Gaussian distributions to approximate arbitrary distributions efficiently. To enhance the difficulty level, we modify the simulated camera distribution to follow a uniform distribution, while initializing the came encoder with Gaussian distribution with higher variances. Remarkably, as shown in \cref{fig:uniform_cam}, the camera encoder which is composed of several Gaussian accurately captures the uniform distribution.

\section{More Visual Results}\label{sec:visual results}
\begin{figure*}[]
	\begin{center}
		\includegraphics[width=0.9\linewidth]{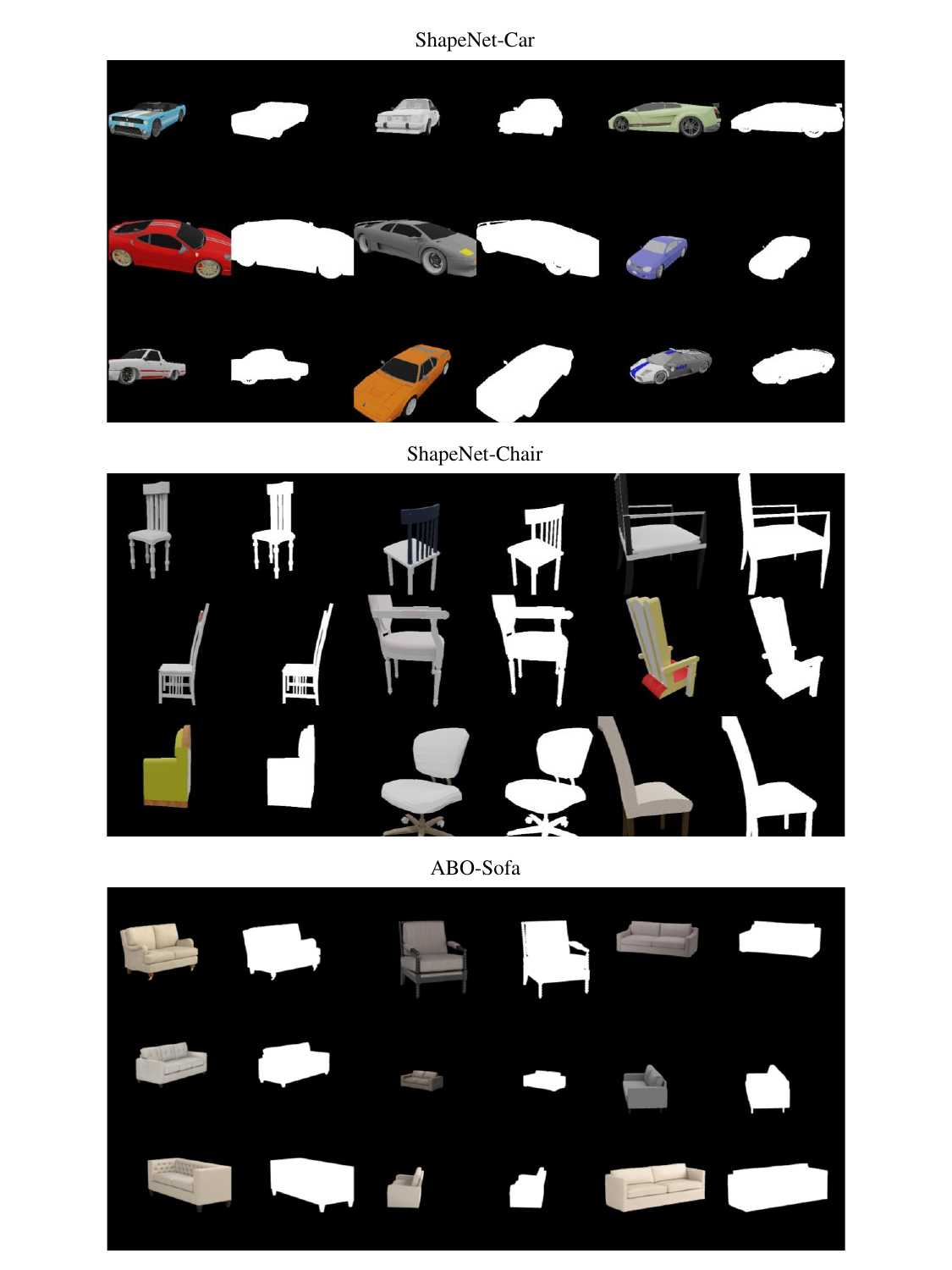}
	\end{center}
    \vspace{-3mm}
	\caption{Snapshots of training images.}
	\label{fig:training}
\end{figure*}

\begin{figure*}[]
	\begin{center}
		\includegraphics[width=0.9\linewidth]{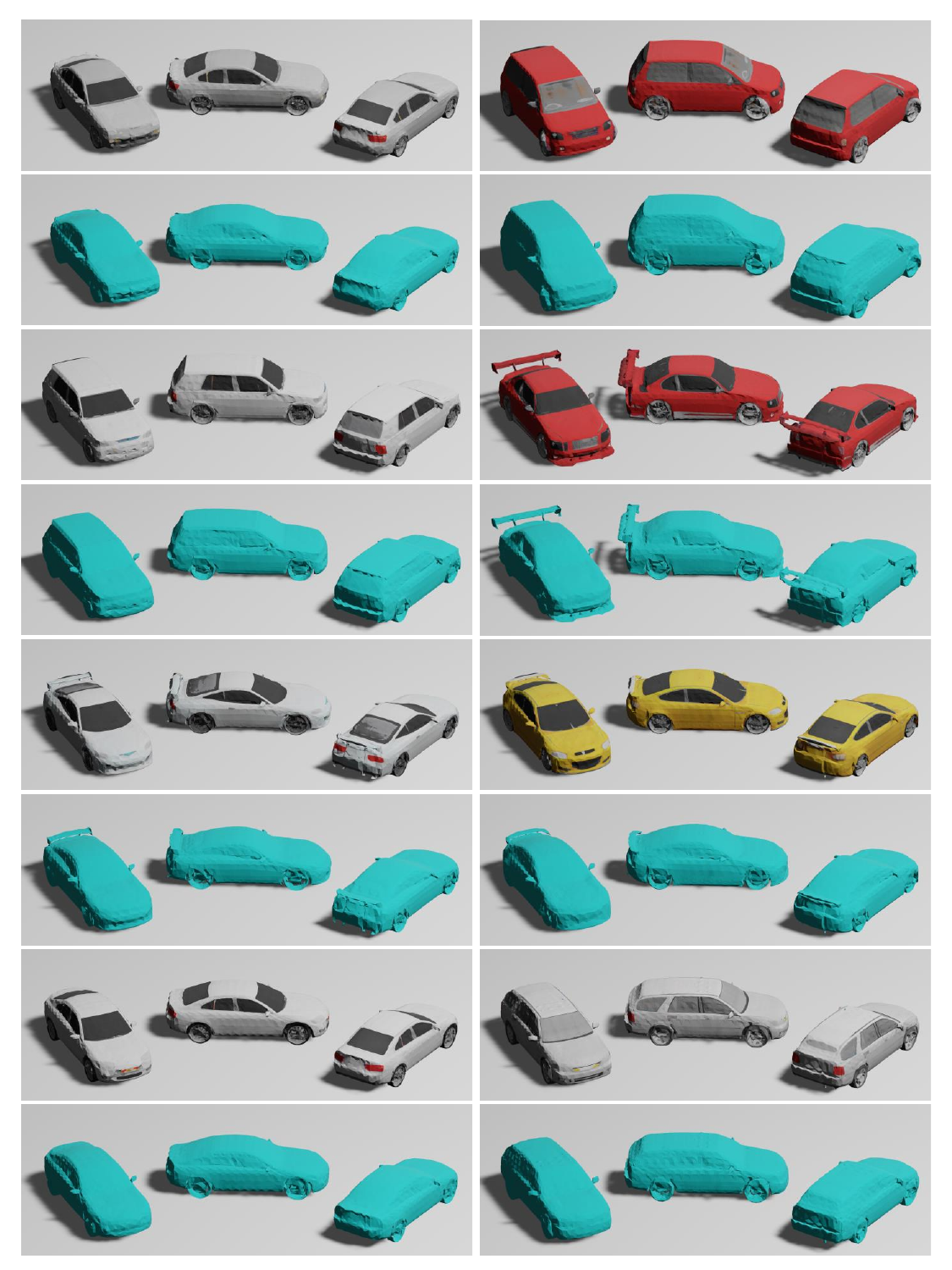}
	\end{center}
    \vspace{-0.5cm}
	\caption{Uncurated qualitative results on ShapeNet-Car.}
	\label{fig:car}
\end{figure*}

\begin{figure*}[]
	\begin{center}
		\includegraphics[width=0.9\linewidth]{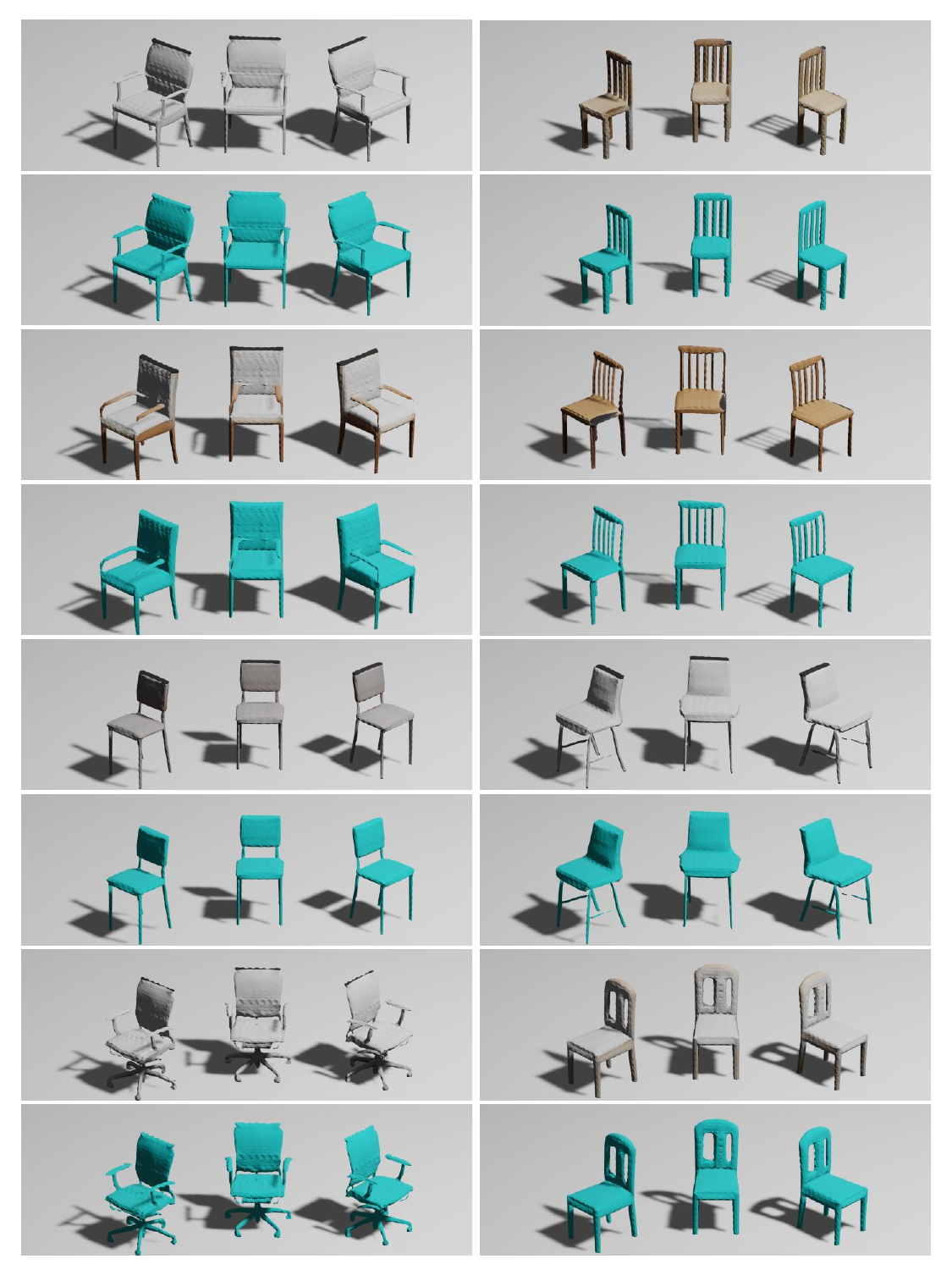}
	\end{center}
    \vspace{-0.5cm}
	\caption{Uncurated qualitative results on ShapeNet-Chair.}
	\label{fig:chair}
\end{figure*}

\begin{figure*}[]
	\begin{center}
		\includegraphics[width=0.9\linewidth]{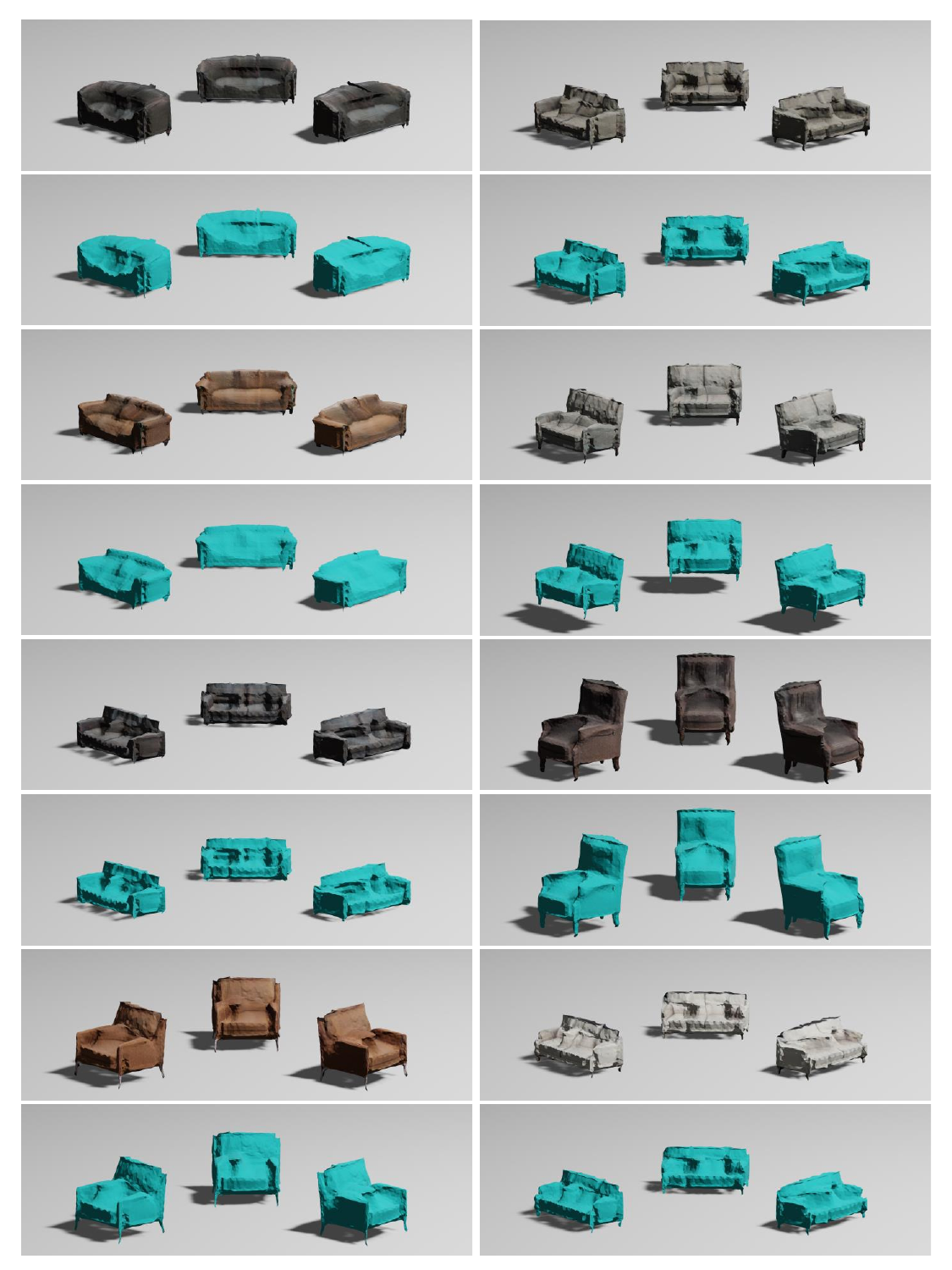}
	\end{center}
    \vspace{-0.5cm}
	\caption{Uncurated qualitative results on ABO-Sofa.}
	\label{fig:sofa}
\end{figure*}

In \cref{fig:training}, we present a snapshot of the unconstrained training data, while \cref{fig:car,fig:chair,fig:sofa} showcase more qualitative results across all datasets. Overall, aside from the limitations discussed in \cref{sec:abo}, \method{} successfully learns high-quality textured 3D shapes from unconstrained 2D images with unknown camera poses.

\end{document}